\documentclass[
	a4paper, % Paper size, use either a4paper or letterpaper
	10pt, % Default font size, can also use 11pt or 12pt, although this is not recommended
	unnumberedsections, % Comment to enable section numbering
	twoside, % Two side traditional mode where headers and footers change between odd and even pages, comment this option to make them fixed
]{LTJournalArticle}

\runninghead{} % A shortened article title to appear in the running head, leave this command empty for no running head

\footertext{\textit{RISC-V Summit Europe, Paris, 12-15th May 2025}} % Text to appear in the footer, leave this command empty for no footer text

\usepackage{balance}
\usepackage[acronym]{glossaries}
\usepackage{siunitx}
\usepackage{booktabs}
\usepackage{multirow}
\usepackage{url}

\title{On-Device Federated Continual Learning on \\ RISC-V-based Ultra-Low-Power SoC for \\ Intelligent Nano-Drone Swarms}

\date{}

\author{%
 	Lars Kr{\"o}ger\textsuperscript{1}, Cristian Cioflan\textsuperscript{2}\thanks{
    Corresponding author: \href{mailto:cioflanc@iis.ee.ethz.ch}{\tt cioflanc@iis.ee.ethz.ch} \\
    This work was partially supported by the Swiss State Secretariat for Education, Research, and Innovation (SERI) under the SwissChips initiative, the EU project dAIEDGE under grant No 101120726, and the ETH-Domain Joint Initiative program (project UrbanTwin).
    }, Victor Kartsch\textsuperscript{2} and Luca Benini\textsuperscript{2,3} \\
    \footnotesize\textsuperscript{\textbf{1}}D-ITET, ETH Zurich; \textsuperscript{\textbf{2}}IIS, ETH Zurich; \textsuperscript{\textbf{3}}DEI, University of Bologna
 }

\renewcommand{\maketitlehookd}{%
	\begin{abstract}
		\noindent RISC-V-based architectures are paving the way for efficient On-Device Learning (ODL) in smart edge devices. 
        When applied across multiple nodes, ODL enables the creation of intelligent sensor networks that preserve data privacy. 
        However, developing ODL-capable, battery-operated embedded platforms presents significant challenges due to constrained computational resources and limited device lifetime, besides intrinsic learning issues such as catastrophic forgetting.        
        We face these challenges by proposing a regularization-based On-Device Federated Continual Learning algorithm tailored for multiple nano-drones performing face recognition tasks.
        We demonstrate our approach on a RISC-V-based 10-core ultra-low-power SoC, optimizing the ODL computational requirements.
        We improve the classification accuracy by 24\% over naive fine-tuning, requiring \qty{178}{\milli \second} per local epoch and \qty{10.5}{\second} per global epoch, demonstrating the effectiveness of the architecture for this task.
	\end{abstract}
}

\begin{document}

\maketitle % Output the title section

%----------------------------------------------------------------------------------------
%	ARTICLE CONTENTS
%----------------------------------------------------------------------------------------

% Generics
\newacronym{lpwan}{LPWAN}{Low-Power Wide Area Network}
\newacronym{lora}{LoRa}{Long Range}
\newacronym{lorawan}{LoRaWAN}{Long Range Wide Area Network}
\newacronym{nbiot}{NB-IoT}{Narrow Band Internet-of-Things}
\newacronym[plural=WANs, firstplural={Wide Area Networks (WANs)}]{wan}{WAN}{Wide Area Network}
\newacronym[plural=WSNs, firstplural={Wireless Sensor Networks (WSNs)}]{wsn}{WSN}{Wireless Sensor Network}
\newacronym{simd}{SIMD}{Single Instruction Multiple Data}
\newacronym{os}{OS}{Operating System}
\newacronym{ble}{BLE}{Bluetooth Low-Energy}
\newacronym{wifi}{Wi-FI}{Wireless Fidelity}
\newacronym[plural=DVS, firstplural={Dynamic Vision Sensors (DVS)}]{dvs}{DVS}{Dynamic Vision Sensor}
\newacronym{ptz}{PTZ}{Pan-Tilt Unit}

% Computer Architecture
\newacronym[plural=FLLs,firstplural=Frequency Locked Loops (FLLs)]{fll}{FLL}{Frequency Locked Loop}
\newacronym{dram}{DRAM}{Dynamic Random Access Memory}
\newacronym{fpu}{FPU}{Floating Point Unit}
\newacronym{dma}{DMA}{Direct Memory Access}
\newacronym[plural=LUTs, firstplural={Lookup Tables (LUTs)}]{lut}{LUT}{Lookup Table}
\newacronym[plural=FPGAs, firstplural={Field Programmable Gate Arrays (FPGAs)}]{fpga}{FPGA}{Field Programmable Gate Array}
\newacronym{dsp}{DSP}{Digital Signal Processing}
\newacronym{mcu}{MCU}{Microcontroller Unit}
\newacronym{spi}{SPI}{Serial Peripheral Interface}
\newacronym{cpi}{CPI}{Camera Parallel Interface}
\newacronym{fifo}{FIFO}{First-In First-Out Queue}
\newacronym{uart}{UART}{Universal Asynchronous Receiver-Transmitter}
\newacronym{raw}{RAW}{Read-After-Write}
\newacronym[plural=ISAs, firstplural={Instruction Set Architectures (ISAs)}]{isa}{ISA}{Instruction Set Architecture}

% Quantization 
\newacronym{ste}{STE}{Straight-Through-Estimator}

\newacronym[plural=PTUs, firstplural={Pan-Tilt Units}]{ptu}{PTU}{Pan-Tilt Unit}
\newacronym{mdf}{MDF}{Medium-density fibreboard}
\newacronym{cvat}{CVAT}{Computer Vision Annotation Tool}
\newacronym{coco}{COCO}{Common Objects in Context}
\newacronym{soa}{SoA}{State of the Art}
\newacronym{sf}{SF}{Sensor Fusion}

% Deep Learning generics
\newacronym{dl}{DL}{Deep Learning}
\newacronym{bn}{BN}{Batch Normalization}
\newacronym{FGSM}{FBK}{Fast Gradient Sign Method}
\newacronym{lr}{LR}{Learning Rate}
\newacronym{sgd}{SGD}{Stochastic Gradient Descent}
\newacronym{gd}{GD}{Gradient Descent}

\newacronym{sta}{STA}{Static Timing Analysis}

\newacronym[plural=GPIOs, firstplural={General Purpose Inupt Outputs (GPIOs)}]{gpio}{GPIO}{General Purpose Input Output}
\newacronym[plural=LDOs, firstplural={Low Dropout Regulators (LDOs)}]{ldo}{LDO}{Low Dropout Regulator}

\newacronym{inq}{INQ}{Incremental Network Quantization}

\newacronym{CV}{CV}{Computer Vision}
\newacronym{EoT}{EoT}{Expectation over Transformation}
\newacronym{RPN}{RPN}{Region Proposal Network}
\newacronym{TV}{TV}{Total Variation}
\newacronym{NPS}{NPS}{Non-Printability Score}
\newacronym{STN}{STN}{Spatial Transformer Network}
\newacronym{MTCNN}{MTCNN}{Multi-Task Convolutional Neural Network}
\newacronym{YOLO}{YOLO}{You Only Look Once}
\newacronym{SSD}{SSD}{Single Shot Detector}
\newacronym{SOTA}{SOTA}{State of the Art}
\newacronym{NMS}{NMS}{Non-Maximum Suppression}
\newacronym{ic}{IC}{Integrated Circuit}
\newacronym{rf}{RF}{Radio Frequency}
\newacronym{tcxo}{TCXO}{Temperature Controlled Crystal Oscillator}
\newacronym{jtag}{JTAG}{Joint Test Action Group industry standard}
\newacronym{swd}{SWD}{Serial Wire Debug}
\newacronym{sdio}{SDIO}{Serial Data Input Output}
% \newacronym{ldo}{LDO}{Linear Dropout Regulator}

\newacronym[plural=PCBs, firstplural={Printed Circuit Boards (PCB)}]{pcb}{PCB}{Printed Circuit Board}
\newacronym[plural=ASICs, firstplural={Application Specific Integrated Circuits}]{asic}{ASIC}{Application Specific Integrated Circuit}

\newacronym[plural=SCMs, firstplural={Standard Cell Memories (SCMs)}]{scm}{SCM}{Standard Cell Memory}
\newacronym{ann}{ANN}{Artificial Neural Networks}
\newacronym{ml}{ML}{Machine Learning}
\newacronym{ai}{AI}{Artificial Intelligence}
\newacronym{iot}{IoT}{Internet of Things}
\newacronym{fft}{FFT}{Fast Fourier Transform}
\newacronym[plural=OCUs, firstplural={Output Channel Compute Units (OCUs)}]{ocu}{OCU}{Output Channel Compute Unit}
\newacronym{alu}{ALU}{Arithmetic Logic Unit}
\newacronym{mac}{MAC}{Multiply-Accumulate}
\newacronym{soc}{SoC}{System-on-Chip}
\newacronym{tcdm}{TCDM}{Tightly-Coupled Data Memory}
\newacronym{pulp}{PULP}{Parallel Ultra Low Power}
\newacronym{ulp}{ULP}{Ultra-Low-Power}
\newacronym{fc}{FC}{Fabric Controller}
\newacronym{ne16}{NE16}{Neural Engine 16-channels}

\newacronym{PGD}{PGD}{Projected Gradient Descend}
\newacronym{CW}{CW}{Carlini-Wagner}
\newacronym{OD}{OD}{Object Detection}

\newacronym{rrf}{RRF}{RADAR Repetition Frequency}
\newacronym{nlp}{NLP}{Natural Language Processing}
\newacronym{qam}{QAM}{Quadrature Amplitude Modulation}
\newacronym{rri}{RRI}{RADAR Repetition Interval}
\newacronym{radar}{RADAR}{Radio Detection and Ranging}
\newacronym{loocv}{LOOCV}{Leave-one-out cross validation}

\newacronym{nas}{NAS}{Neural Architecture Search}
\newacronym{bsp}{BSP}{Board Support Package}
\newacronym{ttn}{TTN}{The Things Network}
\newacronym{wip}{WIP}{Work in Progress}
\newacronym{json}{JSON}{JavaScript Object Notation}
\newacronym{qat}{QAT}{Quantization-Aware Training}

% Metrics
\newacronym{cls}{CLS}{Classification Error}
\newacronym{loc}{LOC}{Localization Error}
\newacronym{bkgd}{BKGD}{Background Error}
\newacronym{roc}{ROC}{Receiver Operating Characteristic}
\newacronym{frr}{FRR}{False Rejection Rate}
\newacronym{eer}{EER}{Equal Error Rate}
\newacronym{snr}{SNR}{Signal-to-Noise Ratio}
\newacronym{flop}{FLOP}{Floating-Point Operation}
\newacronym{fp}{FP}{Floating-Point}
\newacronym{fps}{FPS}{Frames Per Second}

% Datasets
\newacronym{gsc}{GSC}{Google Speech Commands}
\newacronym{mswc}{MSWC}{Multilingual Spoken Words Corpus}
\newacronym{demand}{DEMAND}{Diverse Environments Multichannel Acoustic Noise Database}
\newacronym{kinem}{KINEM}{Keywords In Noisy Environments and Microphones}

% Topologies
\newacronym[plural=BNNs, firstplural={Binary Neural Networks (BNNs)}]{bnn}{BNN}{Binary Neural Network}
\newacronym[plural=NNs, firstplural={Neural Networks}]{nn}{NN}{Neural Network (NNs)}
\newacronym[plural=SNNs, firstplural={Spiking Neural Networks (SNNs)}]{snn}{SNN}{Spiking Neural Network}
\newacronym[plural=DNNs, firstplural={Deep Neural Networks (DNNs)}]{dnn}{DNN}{Deep Neural Network}
\newacronym[plural=TCNs,firstplural=Temporal Convolutional Networks]{tcn}{TCN}{Temporal Convolutional Network}
\newacronym[plural=CNNs,firstplural=Convolutional Neural Networks (CNNs)]{cnn}{CNN}{Convolutional Neural Network}
\newacronym[plural=TNNs,firstplural=Ternarized Neural Networks]{tnn}{TNN}{Ternarized Neural Network}
\newacronym{ds-cnn}{DS-CNN}{Depthwise Separable Convolutional Neural Network}
\newacronym{rnn}{RNN}{Recurrent Neural Network}
% \newacronym{cnn}{CNN}{Convolutional Neural Network}
\newacronym{gcn}{GCN}{Graph Convolutional Network}
\newacronym{mhsa}{MHSA}{Multi-Head Self Attention}
\newacronym{crnn}{CRNN}{Convolutional Recurrent Neural Network}
\newacronym{clca}{CLCA}{Convolutional Linear Cross-Attention}
% \newacronym{mhsa}{MHSA}{Multi-Head Self-attention}
\newacronym{resnet}{ResNet}{Residual Network}

% Audio methods
\newacronym{bf}{BF}{Beamforming}
\newacronym{anc}{ANC}{Active Noise Cancellation}
\newacronym{agc}{AGC}{Automatic Gain Control}
\newacronym{se}{SE}{Speech Enhancement}
\newacronym{mct}{MCT}{Multi-Condition Training}
\newacronym{mcta}{MCTA}{Multi-Condition Training \& Adaptation}
\newacronym{pcen}{PCEN}{Per-Channel Energy Normalization}
\newacronym{mfcc}{MFCC}{Mel-Frequency Cepstral Coefficient}
\newacronym{asr}{ASR}{Automated Speech Recognition}
\newacronym{kws}{KWS}{Keyword Spotting}
\newacronym{odl}{ODL}{On-Device Learning}
\newacronym{odcl}{ODCL}{On-Device Continual Learning}

% Paper specific
\newacronym{ulpm}{ULPM}{Ultra-Low-Power Mode}
\newacronym{lpm}{LPM}{Low-Power Mode}
\newacronym{hpm}{HPM}{High-Performance Mode}
\newacronym{cil}{CIL}{Class Incremental Learning}
\newacronym{cl}{CL}{Continual Learning}
\newacronym{fl}{FL}{Federated Learning}
\newacronym{fcl}{FCL}{Federated Continual Learning}
\newacronym{odfcl}{ODFCL}{On-Device Federated Continual Learning}
\newacronym{mol}{MOL}{Mean Output Loss}
\newacronym{uav}{UAV}{Unmanned Aerial Vehicle}
\newacronym{uwb}{UWB}{Ultra-wideband}

% \vspace{-0.2cm}

\section{Introduction}
\label{sec:introduction}

The advent of energy-efficient RISC-V-based architectures is driving the development of intelligent edge devices, enabling AI-driven applications to continuously learn and autonomously adapt to environmental changes through federated~\gls{odl}.

On-device \gls{cil}~\cite{wibowo2024ofscil} presents challenges in retaining previous knowledge when introducing new classes (catastrophic forgetting).
Instead of storing samples from past classes, regularization-based strategies address this issue by incorporating explicit regularization terms that balance the model's adaptation to old and new classes. 
In a \gls{fl} context, nodes also require sophisticated strategies to distribute knowledge, particularly when only a single node encounters new knowledge domains. 
\gls{fl} algorithms, such as FedProx~\cite{li2020fedprox}, enable distributed learning across multiple devices by facilitating knowledge exchange without sharing raw data -- essential for privacy-sensitive applications.
However, \gls{odfcl} must be optimized to the platform's memory and computational constraints to ensure real-time, energy-efficient performance to meet the application requirements for latency and battery life.

In this work, we address these challenges within the context of \gls{cil} for face recognition, proposing a system for on-device distributed learning on a swarm of nano-drones using FedProx. 
Additionally, we introduce an \gls{odfcl} strategy implementing regularization-based \gls{cil} through \gls{mol}, effectively mitigating catastrophic forgetting. 
Our implementation outperforms naive fine-tuning by 24\%. 
Furthermore, we demonstrate our methodology on a multicore RISC-V \gls{soc}, parallelizing the \gls{dnn} execution on its RISC-V cluster and exploiting its \gls{simd} extensions for the inference of the quantized backbone.  
With a training time of \qty{10.5}{\second} per federated epoch, our \gls{odfcl} approach demonstrates its suitability for extreme edge intelligent devices, made possible by the computational efficiency, flexibility, and scalability of the RISC-V architecture.

\vspace{-0.3cm}

\section{ODFCL System}
\label{sec:system}

We consider a \qty{30}{\kilo parameters}, pretrained, unimodal DSICNet~\cite{brander2023improving} model.
We split it into a frozen, \texttt{int8} part, $M_{F}$, and an \texttt{fp32} part $M_{T}$ (i.e., the classifier and the last convolutional layer) trainable on-device.
The~\gls{dnn} is distributed to $N$ nodes and progressively exposed to $K$ new classes over multiple learning sessions. 
The evaluation is performed on all classes, assessing the model's ability to expand its domain knowledge without forgetting previously learned classes.

Compared to naive fine-tuning, we append a regularization term to the cross-entropy loss.
\gls{mol} minimizes the difference between the activations of the classes learned during the current session and the activations of previously learned classes, while leaving out the output corresponding to the current target class.
This encourages confident predictions and improves the stability of the network.

Each node $n$ locally learns $k$ classes, then the knowledge is aggregated: the parameters $M^{n}_{T}$ are sent by each slave node to an ad hoc master, averaged into a global model, and returned.
Each node also stores a copy of the global model between two synchronization epochs, used to penalize large weight deviations of the local model, thus expanding the loss function with a third regularization term.

We implement our \gls{odfcl} methodology on the GAP9Shield~\cite{muller2024gap9shield}, a Crazyflie\footnote{\url{https://www.bitcraze.io/products/old-products/crazyflie-2-1/}} expansion board featuring GAP9~\cite{gap9}, a multi-core AI-capable \gls{soc}. We exploit its 10 RISC-V cores during inference and training stages and the \gls{simd} extensions for the integerized backbone.
All cores share access to the \qty{128}{\kilo \byte} \gls{tcdm} memory containing activations, gradients, and the weights of the layer currently processed.
The additional on-chip memory (i.e., \qty{1.5}{\mega \byte} RAM and \qty{2}{\mega \byte} flash) is used to store the weights of $M_{F}$, as well as copies of the global and local model $M_{T}$ for regularization.

\begin{figure}[t]
\centering
\includegraphics[width=0.65\columnwidth, trim=0cm 0cm 0cm 0cm, clip]{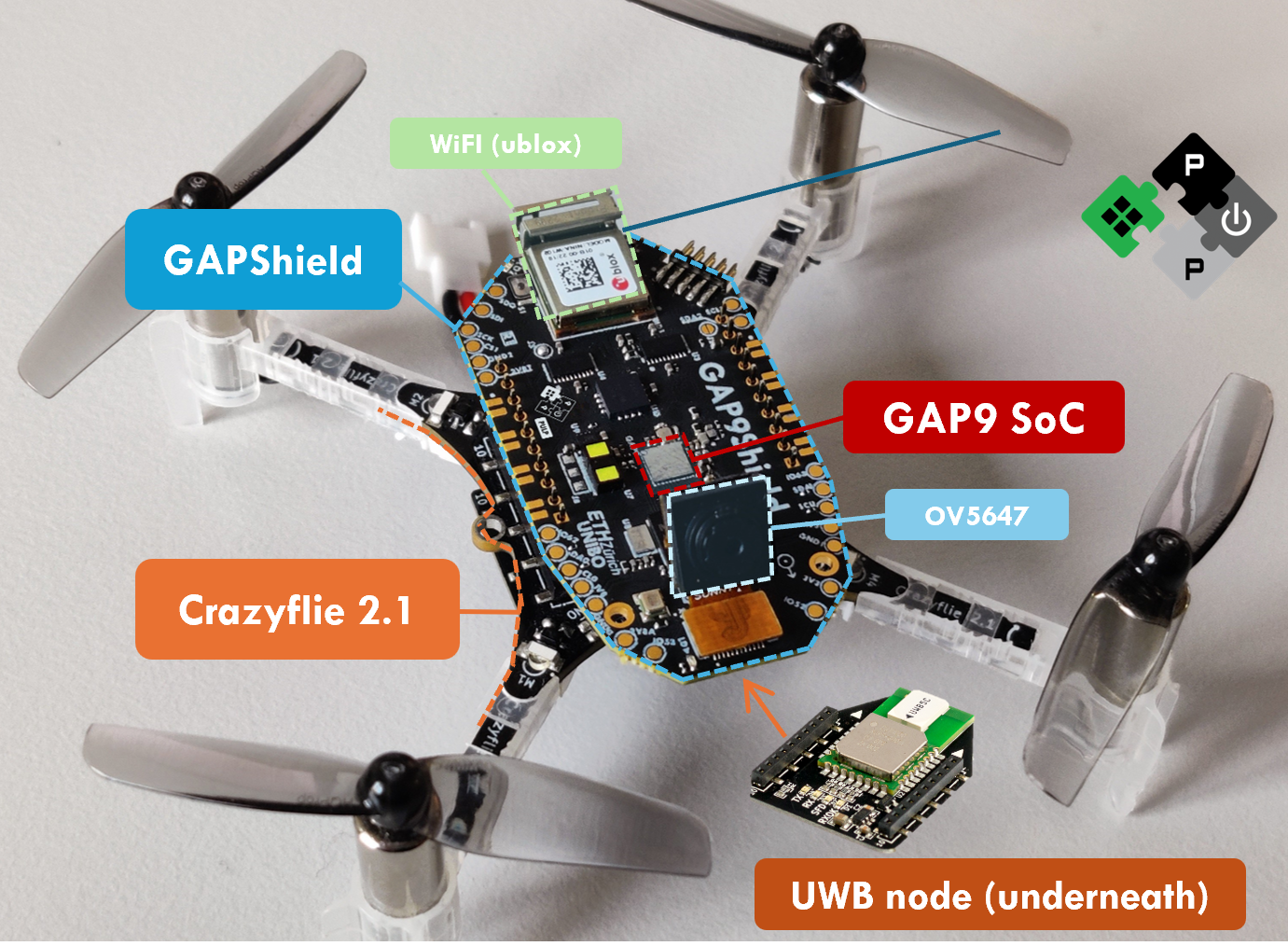}
\caption{GAP9Shield on the Crazyflie 2.1 nano-drone.}
\label{fig:drone}
\vspace{-0.6cm}
\end{figure}

The GAP9Shield mounted on top of the Bitcraze Crazyflie 2.1 nano-drone (i.e., \qty{27}{\gram}, \qty{10}{\centi \metre}) is coupled with the Loco Positioning Deck allowing \gls{uwb} communication during the synchronization stages.
Control and communication are fully managed by the Crazyflie drone using its Cortex-M4 microcontroller. Fig.~\ref{fig:drone} shows the complete platform.

\vspace{-0.3cm}

\section{Results and Conclusions}
\label{sec:results}

We evaluate our \gls{odfcl} system on a split derivative of LFW~\cite{huang2008labeled}, selecting 10 classes, with 28 samples per class for training and 7 for testing.
At T0, DSICNet is trained on four classes, the remaining six distributed in two sessions of three classes each, one per node in a three-node swarm.
We set the regularization factors $\mu = 2$ for \gls{mol} and $\lambda = 3.8$ for FedProx, with model synchronization taking place every training epoch.

Table~\ref{tab:accuracy} depicts the \gls{cil} accuracy of DSICNet, the base accuracy of 65\% indicating the reduced capacity of the model.
Despite the limitations, \gls{odfcl} achieves 46\% on 10 classes, outperforming naive fine-tuning by $2 \times$ thanks to its regularization mechanism.
Compared to an ideal scenario where all data is available beforehand, our system shows a modest 10\% degradation, reducing catastrophic while enabling distributed learning on subsequent tasks.

\begin{table}[t]

\footnotesize

\caption{Accuracy[\%] on Split LFW after each session.}

\label{tab:accuracy}

\centering

\begin{tabular}{lrrr}

Session  & T0 & T1 & T2 \\

\hline

Naive fine-tuning & 65\% & 29\% &  22\% \\ 

\gls{odfcl}    & 65\% & 57\% &  46\% \\

Joint training & 65\% & 54\% & 56\% \\

\vspace{-0.6cm}

\end{tabular}

\end{table}

\begin{table}[t]

\footnotesize

% Cost per sample in the sheet. We have 28 samples per epoch.

\caption{Training cost per local epoch on the GAP9 \gls{soc}.}

\label{tab:power}

\centering

\begin{tabular}{lrr}

Mode & Low-Power & High-Performance  \\
     & Mode (LPM) & Mode (HPM)  \\
     &  $f = \qty{240}{\mega \hertz}$ &  $f = \qty{370}{\mega \hertz}$ \\
     &  $V = \qty{650}{\milli \volt}$ &  $V = \qty{800}{\milli \volt}$ \\

\midrule

% \hline

Latency [ms]  & 178.4   &  117.6   \\ 
Power [mW]    & 24.3  & 53.1   \\
Energy [mJ]  & 4.3 & 6.2  \\

\vspace{-0.9cm}

\end{tabular}

\end{table}

The \texttt{fp32} pointwise convolution represents the largest trainable component, with \qty{29}{\kilo \byte} peak memory requirements during local learning, whereas \qty{24}{\kilo \byte} per node are sent over \gls{uwb} for global aggregation.
As shown in Table~\ref{tab:power}, \qty{118}{\milli \second} are needed for a local training epoch in HPM, as the RISC-V cluster operating at \qty{370}{\mega \hertz} accelerates the training process.
However, the \gls{uwb} weight transmission by the Loco Deck takes \qty{1.7}{\second}, amounting to \qty{10.5}{\second} for a federated epoch on three learning nodes.
In LPM we thus achieve a training latency of \qty{178}{\milli \second} (i.e., \qty{6.4}{\milli \second} per sample), sufficient for 58 local epochs without affecting the system latency, while benefiting from the ultra-low-power architecture at a computational cost per epoch of only \qty{4.3}{\milli \joule}.
These results highlight the benefits of the RISC-V architecture, which allows real-time energy-efficient \gls{odfcl} processing, paving the way for intelligent sensor networks.

Ongoing work considers the deployment of larger \glspl{dnn}, exhausting the remaining \qty{400}{\kilo \byte} of RAM and thus increasing the baseline accuracy.
Furthermore, hybrid \gls{cil} approaches that combine regularization-based strategies with the storage of meaningful latent representations of past classes could further reduce forgetting without increasing the computational cost.

\vspace{-0.3cm}

% \vspace{-0.4cm}

% \input{srcs/conclusion}

% \vspace{-0.4cm}
%----------------------------------------------------------------------------------------
%	 REFERENCES
%----------------------------------------------------------------------------------------

\printbibliography % Output the bibliography

%----------------------------------------------------------------------------------------

\end{document}